\documentclass[11pt,a4paper]{article}
\usepackage{times,latexsym, booktabs}
\usepackage{url}
 \usepackage{graphicx} 
\usepackage[T1]{fontenc}
\usepackage{makecell}
\usepackage{comment}

   \usepackage[acceptedWithA]{tacl2021v1}
\usepackage{xspace,mfirstuc,tabulary}

\newif\iftaclinstructions
\taclinstructionsfalse 
\iftaclinstructions
\renewcommand{\confidential}{}
\renewcommand{\anonsubtext}{(No author info supplied here, for consistency with
TACL-submission anonymization requirements)}
\newcommand{\instr}
\fi

\iftaclpubformat 

\else

\fi

\newcommand{\new}[1]{{#1}}

\title{Scale Can't Overcome Pragmatics: \\The Impact of Reporting Bias on Vision-Language Reasoning}

\author{
  Amita Kamath$^{1,2}$ \hspace{1em}
  Jack Hessel$^3$ \hspace{1em}
  Khyathi Chandu$^4$ \hspace{1em}
  Jena D. Hwang$^5$ \\
  \textbf{Kai-Wei Chang}$^2$ \hspace{1em}
  \textbf{Ranjay Krishna}$^{1,5}$
  \\
  $^1$ University of Washington \hspace{1em}
  $^2$ University of California, Los Angeles \\
  $^3$ Samaya AI \hspace{1em}
  $^4$ Mistral AI \hspace{1em}
  $^5$ Allen Institute for AI
  \\
  \texttt{kamatha@cs.washington.edu} \\
  \url{https://github.com/amitakamath/reporting_bias/}
}

\date{}

\begin{document}
\maketitle

\begin{abstract}
The lack of reasoning capabilities in Vision-Language Models (VLMs) has remained at the forefront of research discourse.
We posit that this behavior stems from a \emph{reporting bias} in their training data.
That is, 
how people communicate about visual content by default omits 
tacit information needed to supervise some types of reasoning; e.g., ``at the game today!'' is a more likely caption than ``a photo of 37 people standing behind a field''.
We investigate
the data 
underlying the popular VLMs OpenCLIP, LLaVA-1.5 and Molmo \new{through the lens of theories from pragmatics}, and find that \new{reporting bias results in}
insufficient representation of four reasoning skills (spatial, temporal, negation, and counting),
\new{despite the corpora being of web-scale, and/or synthetically generated.} 
With a set of curated benchmarks, 
we demonstrate that: 
(i) VLMs perform poorly on the aforementioned types of reasoning suppressed in the training data by reporting bias;
(ii) contrary to popular belief, scaling data size, model size, and to multiple languages does \emph{not} result in emergence of these skills by default; 
but, promisingly, (iii) incorporating annotations specifically collected to obtain tacit information is effective. 
Our findings highlight the need for 
more intentional training data curation methods, rather than counting on scale for 
emergence of reasoning capabilities.

\end{abstract}

\section{Introduction}
\label{sec:introduction}

Research in Vision-Language Models (VLMs) grapples with a paradox: despite impressive performance on standardized benchmarks~\cite{liu2024improved, molmo2024, openai2024gpt4ocard}, models often falter on tasks requiring counting~\cite{paiss2023teaching}, spatial reasoning~\cite{Liu2022VisualSR, kamath-etal-2023-whats} and compositional reasoning~\cite{zhao2022vl, ma2022crepe, yuksekgonul2023when}. 
We hypothesize that these gaps stem from a \emph{reporting bias} in vision-language data. Put simply: when discussing images online,
people systematically omit certain types of information, e.g., spatial prepositions. 
We leverage long-standing bodies of work in linguistics, pragmatics\footnote{Specifically, the Gricean maxims of conversational implicatures \citep{grice1975logic}.}, and cognitive science to identify four types of tacit reasoning systematically omitted by people: \textit{spatial, temporal, counting} and \textit{negations}.

We analyze popular training corpora LAION \citep{laion5b}, LLaVA-1.5 \citep{liu2024improved} and PixMo \citep{molmo2024}, and validate that reporting bias occurs when people write alt-text (as in LAION), when they annotate images with captions (as in PixMo), \textit{and} when captions are synthesized by LLMs (as in LLaVA-1.5). Despite the scale of these datasets, ranging from hundreds of thousands to billions of data points,
instances that operationalize important reasoning remain rare: e.g., we estimate that LAION contains only 0.1\% occurrence of spatial reasoning.

To investigate potential correlations between training data and (a lack of) image-text reasoning skills, we curate evaluation questions that require each of these four types of reasoning.
We evaluate a wide variety of contrastive and generative VLMs on these benchmarks and show that, in line with our hypothesis, 
existing models perform poorly (on average, open-source models fall 54 points behind human performance) unless they are explicitly trained with datasets that require such skills.

Crucially, we find that data+model scaling alone is unlikely to lead to emergent reasoning\footnote{Unlike the success it has shown in perception and recognition tasks \citep{cherti2023reproducible}, which are better represented naturally in training corpora \citep{udandarao2025no}.}---\textit{as the human behaviors underlying the reporting bias do not change with scale}.
Extrapolating scaling performance on our evaluations suggests, e.g.,
that CLIP \citep{RadfordKHRGASAM21} would need to be trained with an intractable 
amount of data or number of model parameters to meet human performance on our benchmarks. 
Adding multilingual diversity to CLIP's training data by translating non-English captions in web-scale corpora to English, as in \citet{nguyen2024multilingual}, also does not improve model performance, showing that the reporting bias is not specific to the English language.

Finally, we study whether annotator instructions can be leveraged to mitigate reporting bias.
We find that for the same underlying images sourced from COCO, instructions from LLaVA and PixMo data collection elicit 2--3 times more instances of counting and spatial reasoning than instructions from COCO. 
Further, with carefully-written instructions we present, negation and temporal reasoning can also be successfully elicited.
The prevalence of reasoning-related information in training data corresponds with improved reasoning capabilities of the corresponding models on our evaluations; 
however, we further verify that our instructions surface sufficient representation of reasoning concepts to improve VLM reasoning in a finetuning setting.
These results show promise to improve model reasoning via intentional data collection, rather than simply scaling.


Our contributions are: (1) revealing the reporting bias in vision-language \new{at even web-scale}, validated with three open-source image-text corpora; (2) re-purposing benchmarks for VLM reasoning and evaluating top-performing contrastive and generative VLMs; (3) revealing that scaling up data, parameters and multilingual diversity do not result in emergent reasoning; and (4) showing that reasoning-aware data collection is possible, and shows promise to improve model reasoning capabilities.
\new{We release our code and data at \url{https://github.com/amitakamath/reporting_bias/}.}

\section{Related Work}
\label{sec:related_work}

\noindent\textbf{Reporting bias} is a well-studied phenomenon in the area of NLP, having presented itself as the ``common sense problem'', e.g., ``people murder'' is a more likely bigram than ``people breathe'' in text\footnote{That people breathe is too obvious of a fact to be expressed in writing.}, leading models trained on this text to incorrectly believe that the former action is more likely to occur than the latter \cite{gordon2013reporting, sap-etal-2019-social, shwartz-etal-2020-unsupervised}. This was overcome with the introduction of large-scale commonsense corpora \cite{bosselut-etal-2019-comet, sap2019atomic} to provide models the lacking information. We study this phenomenon in vision-language data, tackling types of reasoning beyond common sense.

In the vision-language domain, \citet{ye2024computer} show that people from different cultures describe different features of the same image when provided the same instructions. \citet{nguyen2024multilingual} further show that by translating non-English captions to English, VLMs' zero-shot classification performance increases. We acknowledge the increased coverage of information by speakers from different languages, and ask the question: are there types of information omitted by \textit{everyone}? 

Several recent works have studied various failure cases of \textbf{VLM reasoning} \cite{ma2022crepe, zhao2022vl, kamath-etal-2023-whats, hao2025can, yan2025multimodalinconsistencyreasoningmmir}.
In response, other work focuses on improving the quality of the training data by re-captioning images \cite{nguyen2023improving, lai2024veclip,BetkerImprovingIG} and collecting proprietary data \cite{openai2024gpt4ocard}. We investigate a possible cause behind these failure cases, and study open-source datasets to determine whether annotators require specific instructions to include data otherwise omitted due to reporting bias.

\citet{cherti2023reproducible} show that the performance of contrastive VLMs improves across several tasks with an increase in scale of model and training data size. However, this has shown to not be the case for reasoning tasks \citep{AlTahan2024UniBenchVR}. In contrast, we investigate a reason \textit{why} this behavior occurs. Further, our benchmarks target specific types of reasoning, and contain primarily real-world images. Additionally, we study both contrastive and generative VLMs. 

\paragraph{Explanatory hypotheses.} The aforementioned works revealing poor VLM reasoning take 
different stances on the cause of the issue, and thus, its solution. 
Some~\citep{yuksekgonul2023when, hsieh2023sugarcrepe, doveh2023dense, doveh2023teaching} hypothesize that 
failures arise from commonly-used contrastive objectives being too easy, 
and introduce hard negatives to each batch. 
Others hypothesize that image-level losses 
are insufficient, e.g.,~\citet{zeng2021multi} 
introduces a hierarchical loss based on regions. 
We complement these works by 
focusing on the relatively under-studied training \textit{data}.



\section{Reporting Bias in Vision-Language Reasoning}


No matter how large corpora become, if they are sourced from captions written largely by humans, they will exhibit the natural patterns and idiosyncracies of
how humans understand and describe images. 
We leverage long-standing theories from linguistics, pragmatics and cognitive science to arrive at hypotheses of reporting bias, identifying types of reasoning under-represented in web-scale corpora. We then test our accuracy by investigating the training datasets of open-source contrastive and generative VLMs.




\subsection{Theory-based Hypotheses of Omitted Types of Reasoning}
\label{sec:theories}

When people communicate, they do not do so in a vacuum. 
Cognitive semantics points out that a variety of sources such as
intent, perspective, and topic shape the words they use~\cite{langacker2015construal, talmy1972semantic}. 
People use contextual cues to be as expressive as required by the context of the discussion.
Moreover, theories in pragmatics tell us that we are organized in how we achieve this:
we abide by a tacit set of co-operative principles that is expected in communication \cite{grice1975logic, goodman2016pragmatic}. 
These topics have been investigated extensively by various efforts in linguistics, cognitive science, and child language acquisition, inter alia.

We posit that such principles of communication
can help explain the reporting bias we observe in multimodal data.
In writing captions, we produce text that best communicates what we observe. Thus, we expect captions to be subject to the same communicative principles that guide much of our utterances. At the same time, however, caption data is produced in a restricted setting that lacks communicative context that would produce the desired expressiveness. Without knowledge like the topic of discussion and limited understanding of who the caption consumers will be, the caption writers have only basic principles and common knowledge to guide their writing. 

Our investigation focuses on whether or not the 
pragmatic contexts underlying the annotation process of
popular image+text pretraining corpora manifests captions that operationalize the expressive cues necessary to train vision-language models to count, to use negations, and to do spatial and temporal reasoning.


\setlength{\tabcolsep}{5pt}
\begin{table*}[t]
\footnotesize
\centering
\setlength{\tabcolsep}{5pt}
\resizebox{0.9\textwidth}{!}{              
    \begin{tabular}{lcccccccc}
    \toprule
       & \multicolumn{2}{c}{\textbf{Spatial}} & \multicolumn{2}{c}{\textbf{Counting}} & \multicolumn{2}{c}{\textbf{Negation}} & \multicolumn{2}{c}{\textbf{Temporal}}\\
      \textbf{Data} & Occurr. & \makecell{Est. True\\Occurr.} & Occurr. & \makecell{Est. True\\Occurr.} &  Occurr. & \makecell{Est. True\\Occurr.} & Occurr. & \makecell{Est. True\\Occurr.} \\[-.5mm]
      \midrule
      LAION-2B             & 0.3 & 0.1 &  8.8 &  1.7 & 0.8 & 0.1 & 0.9 & 0.2 \\
      COCO              & 3.7 & 3.7 & 10.8 & 10.4 & 0.2 & 0.1 & 0.2 & 0.1 \\
      LLAVA-1.5 (train) & 5.8 & 4.7 & 12.4 &  6.0 & 5.2 & 1.4 & 1.7 & 0.6 \\      
      Molmo (train)     & 3.3 & 2.2 & 28.8 & 16.8 & 6.0 & 3.2 & 2.9 & 0.3 \\   [-.5mm]   
      \bottomrule 
    \end{tabular}
       }
\caption{Percentage Occurrences and Estimated True Occurrences of reasoning-related keywords in popular open-source image-text corpora and training datasets of open-source VLMs.}
\label{tab:occurrence}
\end{table*}



\begin{figure*}[h]
    \centering
    \includegraphics[width=0.98\textwidth]{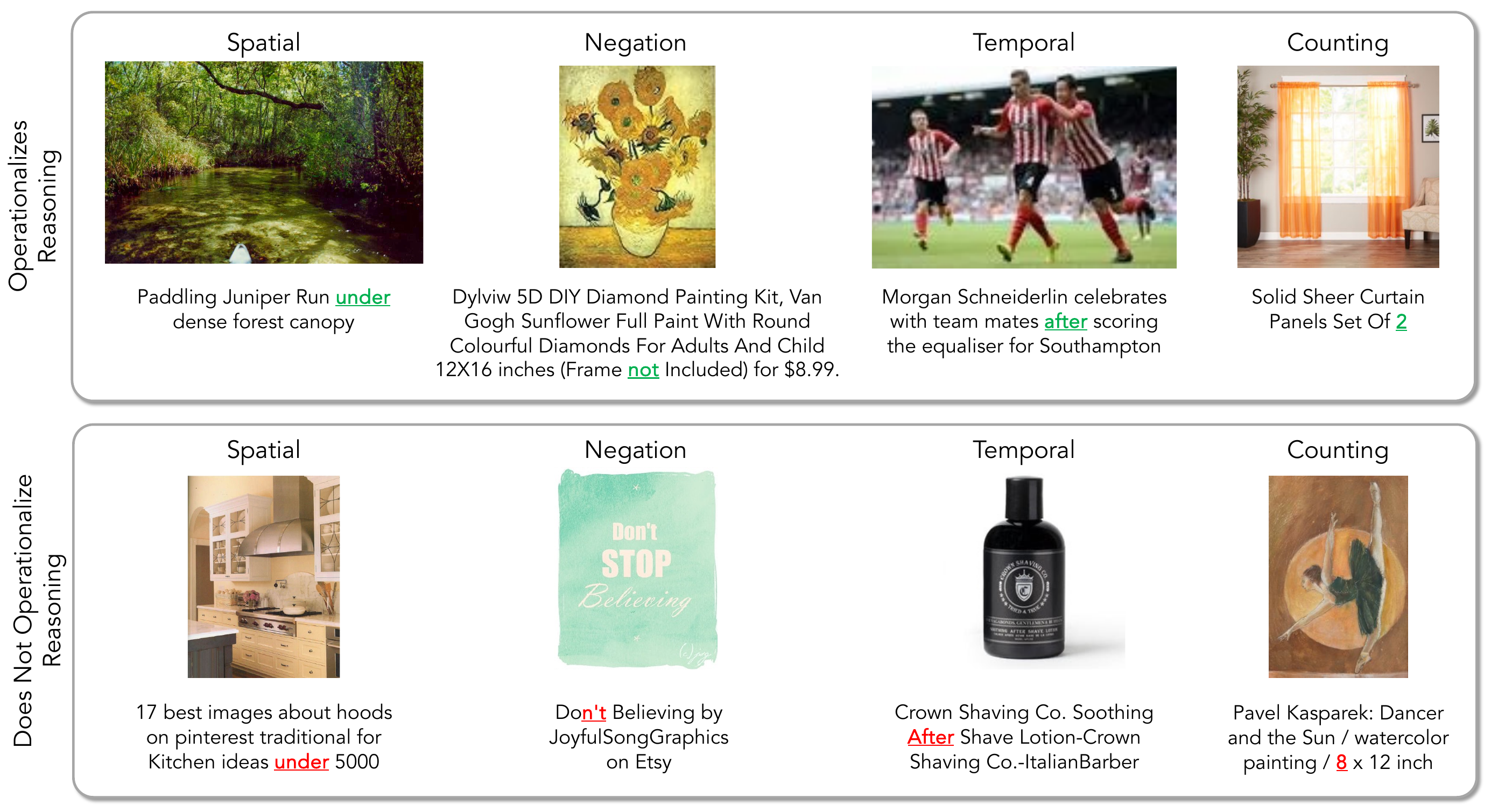}
    \caption{Examples from LAION-2B of data points that contain reasoning-related keywords that do and do not operationalize the reasoning capability itself.}
    \label{fig:qualitative_examples}
\end{figure*}


\paragraph{People tend to omit spatial and temporal language.} 
%
Spatial language such as ``left of'', ``above'' or ``below'' and temporal prepositions such as ``before'' or ``after'' are central to enabling spatial and temporal reasoning respectively. However, unless explicitly directed, people may not naturally produce such language in captioning. 

Pragmatics studies in conversational maxims, known as Gricean Maxims \citep{grice1975logic}, suggest that what information is revealed and how much is revealed is counter-weighed by the expectation to be direct, to be concise and not to misdirect in communication. For example, maxims suggest that even if ``a cat left of a dog'' is a logically accurate description of an image, a person might opt for ``a cat and a dog'' because ``left of'' assigns undue importance to one over the other.
Expressive as it may be, choosing the former caption when there is no explicit reason to do so would be misleading (Maxim of Quality), more information than required (Maxim of Quantity), or would impose a perspective that cannot be justified: whether it is the left of the viewer of the image, or of the subject in the image (Maxim of Manner) \citep{zhang2025do}. 

In the same way, given an image of a boy throwing a ball, writing ``and after, the ball will fall'' would allow for temporal reasoning for a model, but such captions are likely to be avoided because they are too obvious (Maxim of Quantity) or due to insufficient knowledge or evidence about the described event (Maxim of Quality).


Even when spatial preposition use may be merited, studies in cognitive linguistics suggest that captioning may be limited by the existence of default relationships, which we only overlook when the situation calls for it \citep{talmy1972semantic}. For example, when grounding one object (a Figure) with respect to another (a Ground), humans will naturally choose the smaller and easier to move entity as the Figure (e.g., ``a poster above a bed'' is more likely than ``a bed under a poster''). If they are equally sized and movable, we will disprefer the use of spatial language without ulterior reasons.

Such ulterior reasons or perspectives, as theories in linguistics suggest, are provided by discourse mechanisms like the Question under Discussion (QUD)---the implicit or explicitly stated question being addressed in a discourse \cite{von1989referential}. Something as simple as knowing that the image being captioned is a shot of someone's newly adopted cat (given the picture of a dog and a cat) would provide perspectives on how to frame a caption: what to (de)emphasize, what to focus on, or simply, what to talk about \new{(Maxim of Relevance)}. This is not a natural artifact of a restricted annotation setting. Specific prompting (c.f. Section \ref{sec:annotator_instructions}) such as 
``focus on the cat'' or 
``the cat was just adopted'' would be necessary to provide an actionable QUD to trigger the temporal or spatial language we want represented in the data.


\vspace{-0.5em}

\paragraph{People tend to omit counting.} 

Why people may omit object counts in image captions is explained by the expectation that a speaker should maximize the information conveyed while keeping the statement brief (Maxim of Quantity; Rational Speech Act \cite{frank2012predicting, goodman2016pragmatic}). The informational value added by ``six cats'' compared to ``a group of cats'' is negligible without further context, while requiring more effort on the speaker's part (counting the objects).
Moreover, since there are very few contexts in which the listener cares whether there were exactly ``six cats'' compared to ``a group of cats'', i.e., it is rarely the QUD, there is no need for the writer to assume it (Maxim of Relevance). 
\vspace{-0.5em}

\paragraph{People tend to omit negations.} 
Intuitively, there is no rational reason for a person to write ``there are no parrots'' given a picture of a dog and a cat without further context. Much like counting, it would provide more information than necessary to describe the image (Maxim of Quantity; Rational Speech Act) and assign importance when none is merited (Maxim of Quality). 
Additionally, concepts of sentence processing related to psycholinguistics and child language acquisition \cite{tian2016dynamic, pea1978development} suggest that negations are more costly and slower to process than positive statements, and are thus not preferred.


\subsection{Testing Hypotheses in Open-Source Image-Text Corpora}
\label{sec:counts}
In this section, we estimate the frequency of the aforementioned types of reasoning in popular open-source image-text corpora, to test our hypothesis that they occur rarely. We study the training data for OpenCLIP \citep{cherti2023reproducible}, LLaVA-1.5 \citep{liu2024improved} and Molmo \citep{molmo2024}. Where OpenCLIP is only trained on LAION \citep{laion5b}, LLaVA-1.5 and Molmo are additionally trained on open-source academic datasets. We combine the text from all constituent datasets to run this study, taking sampling rates into account as well.

To perform this study, we list keywords corresponding to each type of reasoning, e.g., to study the prevalence of spatial language, we search for the keyword ``right of'' (among other prepositions, c.f. Appendix). While this includes false positives (``right of way''), it loosely upper bounds the prevalence of the spatial relation in the dataset. 
For each keyword, we perform a string search in the listed corpora and show the percentage occurrence of the strings in Table \ref{tab:occurrence} (Occurrence). 

We then sample 100 data points corresponding to each type of reasoning in each corpus and manually calculate the number of data points in which the reasoning is truly represented and visible in the image, i.e., the true positive rate. We calculate a rough estimate of the true number of occurrences of that type of reasoning in the corpus (Estimated True Occurrence in Table \ref{tab:occurrence}). Examples of data points that contain keywords and do or do not operationalize reasoning are shown in Figure \ref{fig:qualitative_examples}. 

As seen in Table \ref{tab:occurrence}, the types of reasoning we study are indeed infrequent even in large-scale corpora, verifying our hypotheses from Section \ref{sec:theories}; e.g., all spatial prepositions we study form a combined estimate of only 0.1\% of LAION. 

\paragraph{\new{Comparison to higher-frequency concepts.}}
While some of the aforementioned frequencies may sound sufficient to learn a concept, they fall far behind when compared to the frequency of more high-occurrence words; e.g., the word ``black'' alone occurs in 3.2\% of LAION captions, and ``white'' in 3\%. 
These concepts are also easier to learn, as they tend to be directly visible in the image (e.g., ``a pair of black shoes'').
This puts the small size of the earlier counts into context. 

\paragraph{\new{Comparison to concepts that compose similarly.}}
Concepts such as colors are also easier to learn as they compose similarly (``a pair of black shoes'' is approximately the same color as ``a black horse'') \citep{saini2022disentangling}. As such, even if some object-attribute pairs are unlikely to occur due to reporting bias in text (e.g., ``yellow banana'' is a less likely bigram than ``green banana''), vision-language models are able to decompose them into their high-frequency components and perform well \citep{paik-etal-2021-world}. 
We thus disregard such cases in our work, and focus on \textit{reasoning} concepts omitted due to reporting bias, which are substantially more difficult to learn as they do \textit{not} compose similarly (``before a game'' looks very different from ``before an exam''). 
It is thus unsurprisingly challenging to learn, e.g., temporal reasoning, from only 0.2\% of the data.

\section{Benchmarks}
\label{sec:benchmarks}
\begin{figure*}[h]
    \centering
    \includegraphics[width=\textwidth]{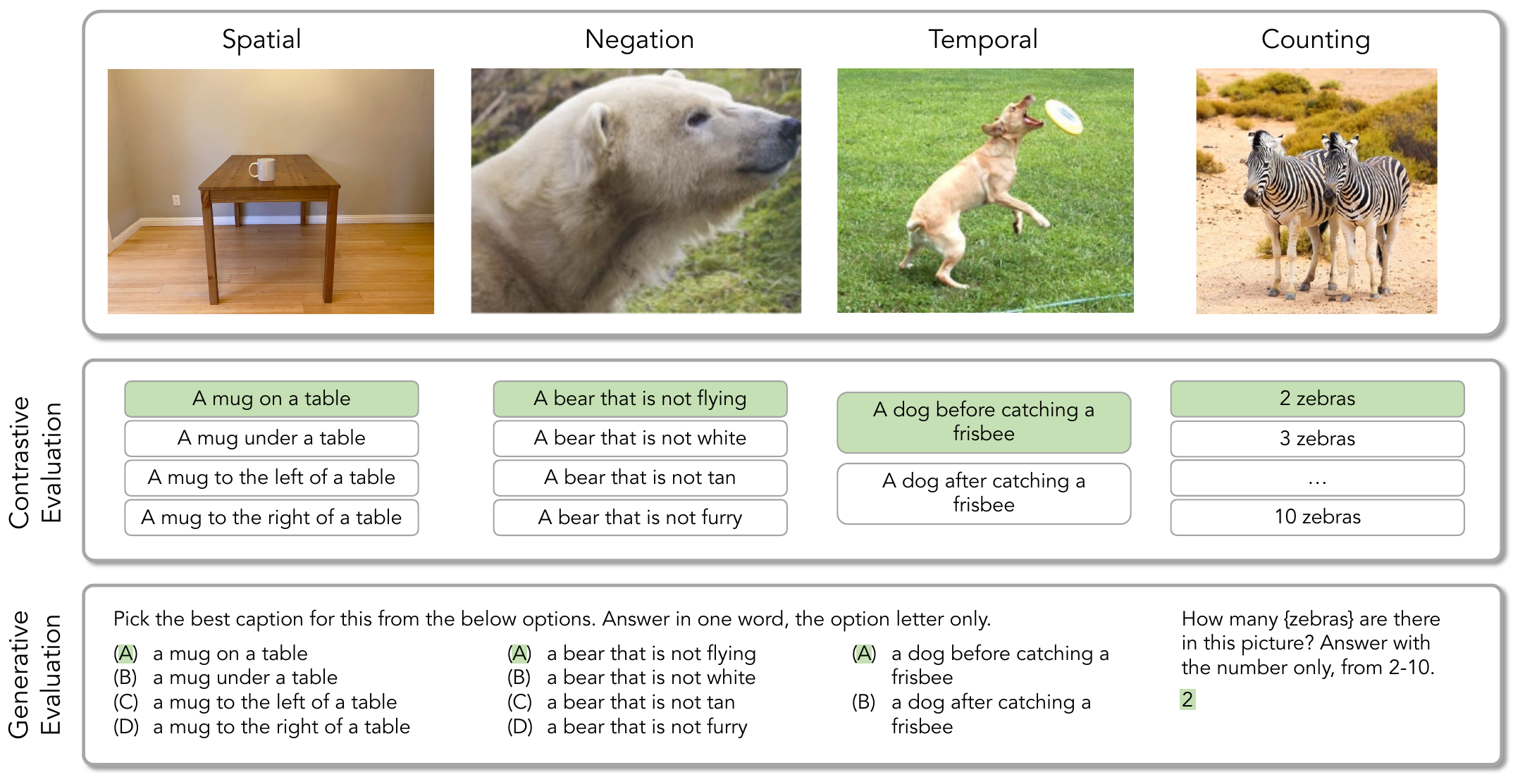}
    \caption{Examples from our four benchmarks for contrastive and generative evaluations. The generative evaluation is in MCQ format but for counting, for which a free form output with a given range yielded higher scores.}
    \label{fig:datasets}
\end{figure*}

Although we have shown that web-scale corpora do not contain significant representation of language related to the types of reasoning we study, models trained on such data could still perform well: they may not require a significant amount of data to learn a skill; or
our keyword-based corpus exploration could have missed relevant data.

We evaluate using four benchmarks across contrastive and generative VLMs, modified from existing benchmarks to suit our needs.
All benchmarks require models to undertake a multiple-choice caption task given an image, as shown in Figure \ref{fig:datasets}. In the case of contrastive VLMs, e.g., CLIP, we take the highest scoring match directly; in the case of generative VLMs, e.g., Molmo, we provide all options at once in a QA-style prompt (except for counting, as we discuss below). 

\paragraph{Spatial reasoning.}
We use Subset A of the What'sUp benchmark \citep{kamath-etal-2023-whats}, targeting four spatial relations: \textit{on}, \textit{under}, \textit{left of} and \textit{right of}. 
The images consist of two basic household objects in a spatial relation to each other, with no distractors. The dataset has 412 data points, perfectly balanced between the four prepositions.

\paragraph{Counting.}
We use a simplified version of CountBench \citep{paiss2023teaching}: originally consisting of captions from LAION that operationalize counting 
(e.g., ``background photo of three light bulbs''), we convert to a simplified format 
by manually reducing each caption to \{count\}\{objects\} (e.g., ``3 light bulbs'') and adding alternate captions for each other count within 2--10. We conduct this modification to ensure the benchmark only evaluates the skill of counting, as well as to avoid data contamination for models that have seen the exact image-text pairs from CountBench in LAION during training. For generative VLMs, we find that all models we evaluate perform better when answering the question directly with a number within a given range, compared to our multiple-choice QA format, and thus report results from the former format for this task alone, as shown in Figure \ref{fig:datasets}.
The dataset contains 507 instances, approximately balanced across counts.\footnote{Several images from LAION are no longer available as of 07/2025, thus resulting in the slight imbalance.}

\paragraph{Negations.}
We re-purpose the VAW benchmark \cite{pham2021learning}, which contains both positive and negative attributes for an object within a given bounding box. We crop the bounding box (discarding those of insufficient size), then write the templated caption ``a photo of a [object name] that is \textbf{not} [attribute]'' with three positive attributes and one true negative attribute, resulting in one correct caption per image. We generate 800 such data points, going through them manually to remove ambiguous attributes such as ``large''. 

\paragraph{Temporal reasoning.}
We use the temporal relations subset of ControlledImCaps \cite{kamath2023text}, which contains pairs of images with corresponding captions: one ``before'' an event, and one ``after''---defining temporal reasoning based on the dynamic context reasoning in VisualCOMET \cite{park2020visualcomet}. We reformat the data to pose a task in which each instance contains one image with two caption options, obtaining 200 data points balanced between ``before'' and ``after''.

\section{Experiments and Results}
\label{sec:experiments}
We evaluate popular contrastive and generative VLMs at various scales of model+data size on our benchmarks to ascertain their reasoning capabilities \new{on types of reasoning less-represented in corpora due to reporting bias}.
We then study the effect on contrastive model performance of scaling both the model parameter size and the training data size, as well as the effect of adding multilingual diversity to the training data. 
Finally, we discuss the performance of popular closed-data and closed-source VLMs on our benchmarks. 

\subsection{Models}
\label{sec:models}
\paragraph{Contrastive VLMs.} 
We consider OpenCLIP \citep{cherti2023reproducible} models of different sizes: ViT-B/32, ViT-B/16, ViT-L/14, ViT-g/14, and ViT-H/14, as well as OpenCLIP ViT-B/32 trained with multilingual diversity, i.e, with non-English captions translated to English added to the data \citep{nguyen2024multilingual}. 

\paragraph{Generative VLMs.} 
We consider two generative VLMs trained on the open-source data examined in Section \ref{sec:counts}:
LLaVA-1.5 \citep{liu2024improved} and Molmo \citep{molmo2024}. We further evaluate several generative VLMs with mixed- or closed-source training data: Qwen-VL \citep{bai2023qwen}, Qwen2-VL \citep{wang2024qwen2}, 
LLaVA-1.6-Mistral \citep{liu2024llavanext}, 
GPT4o and o1 \citep{openai2024gpt4ocard}, Gemini-1.5 Flash and -1.5 Pro \cite{team2024gemini}, and Claude-3 Haiku and -3.5 Sonnet \citep{claude3}. 

\subsection{Results}
\label{sec:results}

\setlength{\tabcolsep}{5pt}
\begin{table}[t]
\centering
\resizebox{\columnwidth}{!}{  

            \begin{tabular}{llcccc}
            \toprule
              &Model & Spatial & Negation & Counting & Temporal\\
              \midrule
              &CLIP ViT-B/32& 30.6 & 11.5 & 43.4 & 58.5 \\
              & \hspace{10px} + ML Div.& 27.4 & 15.5 & 23.3 & 51.5 \\
              &CLIP ViT-B/16& 27.7 & 12.7 & 48.1 & 55.0 \\
              \textcolor{gray}{(a)}&CLIP ViT-L/14& 28.4 & 12.3 & 64.1 & 52.0 \\
              &CLIP ViT-g/14& 28.4 & 12.7 & 59.0 & 52.0 \\
             &CLIP ViT-H/14& 26.0 & 13.2 & 60.0 & 59.0 \\
              \midrule
              &LLAVA-1.5-7B& 37.6 & 33.4 & 47.3 & 72.5 \\
              &LLAVA-1.5-13B& 61.7 & 28.4 & 48.9 & 74.5 \\
             \textcolor{gray}{(b)}&Molmo 7B-O& 75.5 & 38.4 & 77.5 & 78.0 \\
              &Molmo 7B-D& 87.6 & 41.3 & 83.8 & 80.5 \\
              \midrule
              &LLAVA-1.6-m7B& 60.0 & 40.6 & 52.9 & 70.0 \\
              &QwenVL 7B-Chat& 47.1 & 24.2 & 84.6 & 67.5 \\
              &Qwen2VL 7B-Inst.& 98.3 & 56.1 & 85.8 & 84.0 \\
             &GPT4o& 91.5 & 22.2 & 90.9 & 95.0 \\
              &GPT o1& 97.6 & 64.7 & 88.2 & 97.0 \\
              \textcolor{gray}{(c)} &Gemini 1.5-Flash& 98.5 & 46.4 & 84.6 & 81.5 \\
              &Gemini 1.5-Pro& 92.0 & 49.0 & 87.8 & 85.0 \\
              &Claude-3 Haiku& 65.5 & 28.9 & 83.4 & 70.0 \\
              &Claude-3.5 Sonnet& 95.4 & 42.0 & 92.3 & 83.5 \\
              \midrule
              &Random Chance & 25.0 & 25.0 & 11.1 & 50.0\\
              &Human Estimate & 100 & 100 & 100 & 100\\
              \bottomrule \\
            \end{tabular}
        }
\caption{Results on our benchmarks of: (a) Contrastive VLMs, (b) Open-Source Generative VLMs, (c) Closed-Data Generative VLMs. All models fall far behind human performance on multiple types of reasoning.}
\label{tab:main_results}
\end{table}


\paragraph{Contrastive VLMs.} 
Table \ref{tab:main_results}(a) shows the performance of OpenCLIP models on our benchmarks. 
The contrastive VLMs score slightly above random chance on spatial reasoning and temporal reasoning, but score far less than random chance on negations. We find that CLIP tends to ignore negations, scoring the inverse of their attribute detection performance (c.f. Appendix). The models perform fairly well on counting, although it is worth noting that the counting benchmark was initially sourced from OpenCLIP training data. \new{Performing poorly on spatial, negation and temporal aligns with the extremely low occurrence of keywords corresponding to those types of reasoning in LAION, as shown in Table \ref{tab:occurrence}, as does the higher performance of CLIP on counting and the higher occurrence of counting keywords in LAION.}

\paragraph{Generative VLMs.} 
Table \ref{tab:main_results}(b) shows the performance of open-source generative VLMs.
The generative models outperform the contrastive models on average, but fall far behind human performance across all tasks, especially negation. Scaling up LLaVA-1.5 significantly improves spatial reasoning performance, but no other type of reasoning. \new{In the context of keyword occurrences, improvements in reasoning over contrastive models align with increases in corresponding keywords in respective training data in Table \ref{tab:occurrence}.}

\paragraph{Human Performance.} All models, particularly those trained on open-source data, fall far behind human performance across tasks. The human performance results, estimated by collecting annotations from pairs of expert annotators, emphasize that these types of reasoning are trivial to humans.

\begin{figure*}[h]
    \centering
    \includegraphics[width=0.99\textwidth]{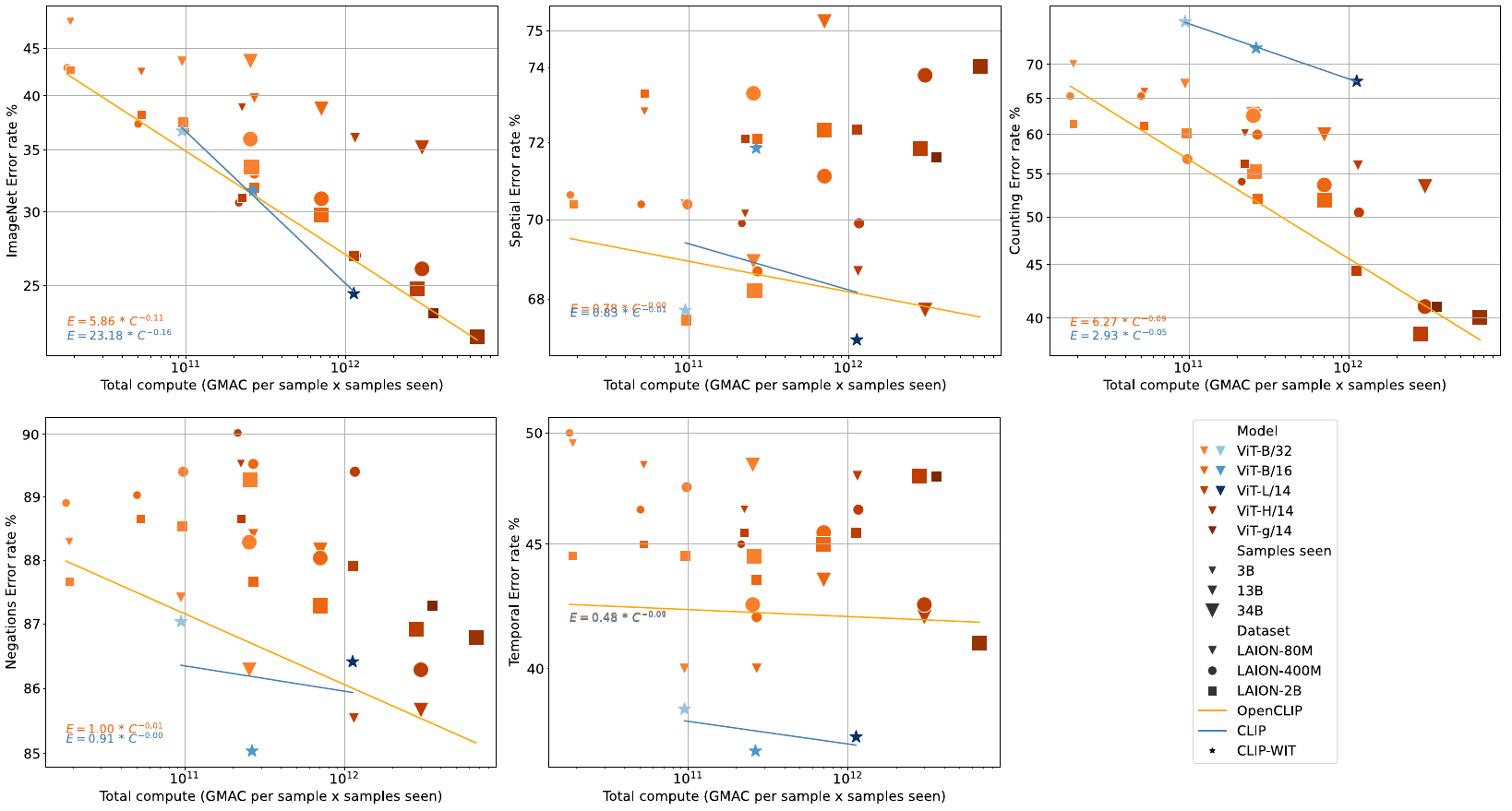}
    \caption{Scaling laws for OpenCLIP models on ImageNet (top left) compared to our benchmarks on spatial, counting, negation and temporal tasks respectively. Note the log-log plots and differing $y$ axes across graphs.}
    \label{fig:scaling_laws}
\end{figure*}

\subsection{Scaling Laws}
\label{sec:scaling_laws}
In this section, we evaluate the aforementioned OpenCLIP models with different training data sizes (LAION-80M, LAION-400M, LAION-2B) and number of data points seen during training (3B, 13B, 34B), obtaining 32 models in total. Each of these is evaluated on our benchmarks to obtain scaling laws, as in \citet{cherti2023reproducible}. The resulting graphs are shown in Figure \ref{fig:scaling_laws}. In contrast to CLIP behavior on pure perception tasks such as ImageNet \cite{Deng2009ImageNetAL}, where the loss drops steeply with an increase in data and/or parameter scale \cite{cherti2023reproducible}, on our benchmarks we see different patterns: on spatial reasoning, the scaling law struggles to fit the data points, but it is clear that the loss does not drop with an increase in compute; on counting, increasing compute does seem to help, but noting the log scale, the amount of compute would need to be several orders of magnitude higher to reach human performance (at 0\% loss); on negation, increasing compute helps very slightly, but the loss remains very high ($\sim$87\%), and an intractable amount of compute would be needed to reach human performance (at 0\% loss); and on temporal reasoning, increasing compute does not improve performance. 

Note that the prevalence of counting data far surpasses that of negations, temporal or spatial relations (c.f. Section \ref{sec:counts}), explaining its relatively high performance---although the frequency is still low on average compared to popular attributes, and the model performance is far behind human performance, which is 100\%. 

When we disentangle model scale from data scale, we see equivalent trends.
From this, we infer that neither scaling up the model size, the training data size, nor both, improves model performance beyond what is seen in Figure \ref{fig:scaling_laws}---proving that the underlying problem of reporting bias cannot be mitigated with scale alone, as an intractable amount of compute in the form of training data and/or model parameters would be needed to reach human performance on these benchmarks.


\subsection{Adding Captions from Other Languages}
\label{sec:multilingual}
\citet{nguyen2024multilingual} showed that adding multilingual diversity to the training data of contrastive VLMs by translating non-English web-scraped alt-text to English can significantly improve their performance on classification tasks; this work was rooted in \citet{ye2024computer}, which highlighted the difference in semantic content in images discussed when people using different languages captioned the same image. We ask: is leveraging this multilingual diversity sufficient to circumvent the reporting bias seen in image-text corpora? To study this, we evaluate the OpenCLIP ViT-B/32 model from \citet{nguyen2024multilingual} on our benchmarks. As seen in Table \ref{tab:main_results}, this model actually underperforms the OpenCLIP ViT-B/32 model trained on LAION English captions alone---showing that these types of reasoning are omitted by \textit{all} speakers.


\subsection{Closed-Data Generative Models}
\label{sec:closed_data}
Top-performing closed-data and closed-source models perform well on our benchmarks, although they still fall behind human performance, especially on negation and temporal reasoning. As the details behind the data collection and training are not public, it is difficult to draw inferences from these results; however, the importance of data quality in addition to scale is clear from efforts invested in data collection \cite{openai2024gpt4ocard}.
\section{\new{Leveraging Annotator Instructions Can Mitigate Reporting Bias}}
\label{sec:annotator_instructions}

Annotator instructions have a significant impact on reporting bias. While people may default to certain behaviors while communicating about images, they can be specifically instructed to discuss certain concepts that they would otherwise tend to omit. 
We study the correlation between annotator instructions and reporting bias in existing datasets, then leverage annotator instructions in a user study to mitigate reporting bias. We verify that our mitigation method surfaces sufficient representation of reasoning-related concepts to improve VLM reasoning in a finetuning setting, and close with broader implications for data collection methods for vision-language corpora.





\subsection{Impact of Annotator Instructions on Reporting Bias in Existing Datasets}
\label{sec:annotator_in_existing}
\new{We study the effect of annotator instructions provided during collection of existing datasets on reporting bias, i.e., on the occurrence of reasoning-related keywords in captions, shown in Table \ref{tab:occurrence}. Annotator instructions are provided in Appendix.}

We first study a dataset where no annotator instructions are provided at all: LAION \citep{laion5b}, which was scraped from alt-text fields of images on the internet. We see from Table \ref{tab:occurrence} that LAION has low representation across all four types of reasoning we study. \new{This reflects the default behavior of people captioning images: when not given any specific instructions, people's tendency to under-report available information, i.e., reporting bias, is seen clearly.}

We next look at COCO's \citep{chen2015microsoftcococaptionsdata} crowdsourced captions, where 
annotators were given instructions, but no specific prompting that would engage them in reasoning.
In fact, they were explicitly instructed to ``\textit{not} describe things that might have happened in the future or past''. 
Accordingly, we observe in Table \ref{tab:occurrence} that the instructions lead to an even lower occurrence of temporal reasoning in COCO as compared to the non-existent instructions of LAION. 
Interestingly, however, we see that the prevalence of spatial language and counting in COCO is higher than that of LAION. Having temporal reasoning restricted, annotators may have turned to focus more closely on describing the objects in the image. 

For LLaVA-1.5 \citep{liu2024improved},
the instructions required discussion of
``object counts'' and ``relative positions between objects'', among other non-reasoning-related instructions. This leads to higher occurrences of both counting and spatial reasoning than in COCO. However, it is worth pointing out that their estimated true occurrences are \textit{not} higher than that of COCO. This may be explained by LLaVA's use of GPT-4 as annotator for instruction tuning data. Our analysis shows that many of the false positives are in fact spurious descriptions that incorrectly use counts and spatial language (e.g., a ``left of'' that is actually a ''right of''), which is consistent with GPT-4's weaknesses in reasoning. Had human annotators been employed, we expect to have observed higher true occurrences of counting and spatial language.



Finally, for Molmo's \citep{molmo2024} pretraining data, the annotators were instructed to discuss ``objects and their counts'' and ``positions of the objects'', among other non-reasoning-related instructions. Molmo's training data includes PixMo as well as other academic datasets, e.g. TallyQA \citep{acharya2019tallyqa} and VQAv2 \citep{goyal2017making}. 
As seen in Table \ref{tab:occurrence}, specific instructions for counting and spatial leads to increased prevalence of spatial and counting reasoning. Without specific instructions, negations and temporal remaining remain low, as in LLaVA and LAION.
It is important to note that there is additional data in PixMo to assist models with spatial reasoning and counting that is in the form of bounding box coordinates, and as such is not included in the above occurrence estimates. 


\subsection{User Study to Mitigate Reporting Bias with Annotator Instructions}
\label{sec:controlled_study}
\new{From these observations, it is clear that instructing annotators to include a certain type of reasoning \textit{does} result in representation of the same.}
\new{However, these observations are drawn from datasets with different image distributions, in addition to having different annotator instructions.}

To further test our hypothesis, \new{we disentangle the two by} carrying out a controlled study where annotators are given a fixed set of 100 images randomly sampled from COCO and requested to caption them. We provide them with one of four sets of annotator instructions: the original COCO captioning instruction, the LLaVA-1.5 captioning instruction, the PixMo captioning instruction, and instructions we write. We re-format the instructions slightly (e.g., PixMo captions were collected via audio, not text), but we retain the \textit{exact} wording of what annotators were requested to include and not include in the captions. In our own instructions, we ask specifically for all four types of reasoning we study.
All sets of instructions are provided in the Appendix.

\begin{table}[t]
\footnotesize
\centering
\resizebox{\columnwidth}{!}{              
    \begin{tabular}{lcccc}
    \toprule
        Instructions & Spatial & Counting & Negation & Temporal \\
        \midrule
        COCO  & 8 & 23 & 2 & 2 \\  
        LLAVA-1.5  & 17 & 38 & 3 & 0 \\
        PixMo  & 21 & 43 & 12 & 1 \\
        \midrule
        Ours  & 14 & 39 & 52 & 44 \\
      \bottomrule 
    \end{tabular}
       }
\caption{Percentage True Occurrences (manually calculated) of reasoning-related keywords in each set of 100 captions collected with different instructions for the controlled study. 
}
\label{tab:controlled_study}
\end{table}

We use Prolific\footnote{\url{https://www.prolific.com/} [accessed 1/2026]} to collect participants for the study. They were asked to write a caption of at least 8 words (the minimum caption length in COCO), but were encouraged to make the captions as long as needed to include the requested information (which varied based on the instruction set). By not constraining the caption length, we mirror the tendency of people to communicate concisely (Maxim of Quantity). Annotators were paid \$15 per hour of estimated work, with a post-task bonus 
if they spent longer on the task. This allowed us to simulate the concise nature of communication (the annotators did not know they would be paid additionally) while paying annotators fairly.

We then check the 100 written captions for percentage occurrences as in Section \ref{sec:counts}, manually calculating the true positive rate. The results are shown in Table \ref{tab:controlled_study}. When annotators are not asked to include anything specific, as in COCO, they do use some spatial- and counting-related words, but no negation- or temporal-related words. Adding requests for spatial and counting, as in LLaVA-1.5 and PixMo, significantly increases the occurrence of words related to those types of reasoning, but \textit{not} to temporal relations or negations. 
By specifically instructing all four phenomena, as in our instructions, we see that the prevalence of all four types of reasoning increases compared to COCO.

These results show that our findings from Section \ref{sec:annotator_in_existing} hold regardless of image distribution. Additionally, 
\new{if the cause of the initial under-representation of reasoning-related keywords were for a different reason than simply people's tendency to under-report available information (i.e., reporting bias), e.g., if these concepts were objectively difficult to capture in text, then modifying the annotator instructions would not have increased representation of the same. As such, our findings further cement reporting bias as a cause for under-representation of reasoning-related keywords in image-text corpora.}


\subsection{Does Our Method Sufficiently Mitigate Reporting Bias?}
\label{sec:finetuning}
Our experiments show strong evidence that reporting bias causes low occurrence of reasoning-related concepts in training data, which in turn correlates with poor model performance on tasks requiring the corresponding types of reasoning. Having put forward a method to mitigate reporting bias, we now ask: is this sufficient to improve the reasoning capabilities of VLMs?


Answering this question would require the creation of
a large-scale vision-language pretraining dataset collected with our recommended annotator instructions (or via other methods that similarly mitigate reporting bias). Unfortunately, this is beyond our resources. 
Instead, we run finetuning experiments
to obtain some signal about whether our method increases the representation of reasoning-related concepts in the collected data enough to improve VLM reasoning in this setting.

We first curate a dataset with counting data equivalent to the percentage occurrence of counting data we surface with ``our'' annotation instructions in Section \ref{sec:controlled_study}, i.e., 39\%.
To achieve this, we sample a balanced subset of counting data from TallyQA (1000 examples per count between 2--10) and combine it with the corresponding proportion of LLaVA-1.5 instruction tuning data (which we estimate to have 6.9\% counting data\footnote{This is a subset of the LLaVA-1.5 training data studied in Section \ref{sec:counts}, and thus has a slightly different occurrence rate for counting than is reported in Table \ref{tab:occurrence}.}).
In total, we combine 9,000 data points from TallyQA with 17,138 data points from LLaVA-1.5 instruction tuning data, to obtain a corpus of about 26K examples with 39\% occurrence of counting data.

\begin{table}[t]
\footnotesize
\centering
\resizebox{\columnwidth}{!}{              
    \begin{tabular}{lcc}
    \toprule
        Model & \makecell{\% Count.\\Data} & \makecell{Counting\\Perf.} \\
        \midrule
        LLaVA-1.5-13b & 6.0 & 49.8 \\  
        FT on LLaVA-IT (26K) & 6.9 & 50.7 \\
        FT on LLaVA-IT + TallyQA (26K)& 39.0 & \textbf{54.4}\\
      \bottomrule 
    \end{tabular}
       }
\caption{Estimated amount of counting data and Counting performance of LLaVA-1.5-13b and finetuned versions. Our instructions (Section \ref{sec:controlled_study}) elicit 39\% counting data (Table \ref{tab:controlled_study}), which is sufficient to improve model counting over simply increasing the data.
}
\label{tab:finetuning}
\end{table}

We finetune LLaVA-1.5-13b on this data for 1 epoch on 2 L40S GPUs at a batch size of 4 and learning rate of 1e-6, then evaluate the finetuned model on our counting benchmark (Section \ref{sec:benchmarks}). 
To ensure the proportion of counting concepts in the finetuning data is the cause of this gain, we compare to finetuning LLaVA-1.5-13b on a corpus of the same size with less occurrence of counting data, i.e., only LLaVA instruction tuning data.
Our results are shown in Table \ref{tab:finetuning}: finetuning on data with less reporting bias outperforms both the base model, and finetuning on data with reporting bias.

While
finetuning on a distribution of data with less reporting bias, i.e., with higher representation of reasoning-related concepts, unsurprisingly increases model performance on those types of reasoning, 
this experiment shows that: 
(1) these types of reasoning are not precluded by the architecture of these models, which agrees with our results from Table 2 and findings from \citet{paiss2023teaching}, \citet{chen2024spatialvlm}, and \citet{ogezi-shi-2025-spare}, which specifically collect data to improve a certain type of reasoning;
and (2) our annotator instructions surface sufficient representation of reasoning-heavy concepts (here, counting) for the model to improve on the task, i.e., they successfully mitigate reporting bias.

While the latter finding has promising implications for data collection methods, the significant gap remaining between the finetuned model and human performance reported in Table \ref{tab:main_results} highlights room for improvement beyond simple finetuning---underscoring the importance of mitigating reporting bias while collecting large-scale pre-training corpora for VLMs.

\subsection{\new{Implications for Data Collection Methods}}
Our findings 
have strong implications for the use of annotator instructions to prevent reporting bias in future data collection efforts.


\paragraph{Reasoning \textit{can} be elicited from annotators.} We show that all types of reasoning we study \textit{can} be elicited from annotators, \textit{if} they are explicitly asked for the same. In terms of our linguistics study in Section \ref{sec:theories}, by making the Question Under Discussion explicit, we are able to elicit the desired information.

\paragraph{Instructions to prevent reporting bias do not generalize across types of reasoning.}
We observe that instructing annotators to include a specific type of reasoning encourages them to discuss that type of reasoning, but not any other. This emphasizes the need to be intentional with annotator instructions for each type of reasoning, if representation of various types of reasoning is desired.


\paragraph{Reporting bias can't be circumvented with caption length alone.}
We perform a study to determine whether forcing increased caption length as in dense annotation schema (e.g., PixMo \citep{molmo2024} requiring annotators to speak about the image for a full minute) yields reasoning-related information without the need for specific instructions. We find that it increases the occurrence of the types of reasoning people were already predisposed to in the original COCO study, but not of the other types of reasoning. Details are in the Appendix.

\paragraph{Reporting bias does impact model performance.}
As discussed in Section \ref{sec:results}, low occurrence of reasoning-related data in training corresponds with poor model performance on that type of reasoning, and the converse (that increased occurrence corresponds with increased performance) is also true. As such, our findings shed light on a promising method to improve VLM reasoning.

\paragraph{Reporting bias occurs in LLM-synthesized data too!}
LLaVA-1.5 serves as an interesting case study, because a significant amount of the data is synthetically generated with GPT-4. 
With increasing emphasis on scale during VLM training, data synthesis methods are becoming correspondingly more popular \citep{wang2024qwen2, bai2025qwen25, liu2024llavanext}. Our study shows that language models are not immune to reporting bias (they, too, are trained on primarily human-written data), and that
instructions given to LLMs to synthesize data are of the same importance as instructions given to human annotators, in terms of their impact on mitigating reporting bias from the generated data. Further, a solution must be found for the chicken-and-egg problem of VLMs being poor at reasoning due to reporting bias, and in turn, generating low-quality synthetic data for model training.

Altogether, our study makes it clear that future data collection methods must be intentional about ensuring representation of various types of reasoning in corpora despite reporting bias. In other words, annotator instructions (or LLM instructions, in the case of synthetically-generated data) are key to overcoming reporting bias and improving model reasoning capabilities. 
\section{Conclusion and Future Work}
\label{sec:conclusion}
We study the \textit{reporting bias} in vision-language: specifically, the systematic omission of types of information by people captioning images, which then form the image-text corpora popular VLMs are trained on.
By identifying human behaviors rooted in linguistics, pragmatics, and cognitive science, we predict the types of information omitted, verify their lack in public image-text corpora, and show that contrastive and generative VLMs trained on this data perform poorly on the types of reasoning corresponding to the missing information. Further, we reveal the importance of the instructions provided to annotators during data collection, showing that intentional collection shows promise in improving representation of reasoning-related data in training corpora, which could in turn improve reasoning capabilities of VLMs. 

Future research directions include: (1) automating identification of significant gaps in image-text corpora; 
(2) synthesizing high-quality data to fill those gaps; (3) finetuning models on augmented data using different methods;
and (4) eliciting captions that avoid the reporting bias in a more natural way than programmatic augmentation, e.g., with our annotation instructions, or by identifying communicative intents that naturally call for these types of reasoning-related information. 

\section*{\new{Limitations}}
Our work highlights reporting bias in image-text corpora as a key factor behind why VLMs trained on these corpora struggle with types of reasoning basic to humans.
Our first major limitation is that, while we suggest reasoning-aware annotator instructions as a method to circumvent reporting bias, and show its promise in Sections \ref{sec:controlled_study} and \ref{sec:finetuning}, actually generating training data at scale using our proposed method is beyond our resources.
Our second limitation is that we present scaling laws in Section \ref{sec:scaling_laws} as evidence suggesting that simply increasing model or data scale does not improve reasoning capabilities. However, while well-studied in the language, vision-language and vision fields \citep{kaplan2020scaling, cherti2023reproducible, AlTahan2024UniBenchVR}, scaling laws are technically a hypothesis, and could fail to hold at much higher scales \citep{nakkiran2021deep}.


\section*{Acknowledgments}
We would like to thank William Merrill, Ananya Harsh Jha, Nishant Subramani, and the members of the RAIVN lab for helpful discussions and feedback, particularly during the conception of this project. This work was partially supported by the Allen Institute for AI, SRC Jump 2.0 CoCoSys, ONR grant N00014-23-1-2780, DARPA ANSR program FA8750-23-2-0004, ECOLE HR00112390060, and Apple.

\bibliography{tacl2021}
\bibliographystyle{acl_natbib}

\appendix
\clearpage

\section{Appendix}
\subsection{Details about Occurrence of Reasoning-Related Keywords}
\paragraph{Keywords.} The keywords we search for are: (1) ``on top of'', ``under'', ``left of'' and ``right of'' for spatial reasoning; (2) ``before'' and ``after'' for temporal reasoning; (3) ``two''--``ten'' and ``2''--``10'' for counting; and (4) ``not'' and ``n't'' for negations.

\paragraph{Estimating True Positive Rate.} 
We discard keyword occurrences that do not operationalize the types of reasoning we study, e.g., ``jeans under \$25'' does not encourage spatial reasoning.

\paragraph{Discussion about choice of keywords.} While the set of keywords we use for our corpus search is limited, it closely aligns with our evaluation of each type of reasoning, as shown in Figure \ref{fig:datasets}. We sample the data to ensure we do not miss keywords operationalizing each type of reasoning (including synonyms of our keywords).
Some keywords are dropped due to their appearance primarily \textit{not} operationalizing the type of reasoning we study: e.g., ``on'' could be a spatial preposition, but is used overwhelmingly in non-spatial contexts (e.g., ``on sale'', ``on record'', ``on demand'', ``on January 28'', etc.), and calculating ``Estimated True Occurrence'' as in Table \ref{tab:occurrence} with a small sample showed close-to-no spatial contexts.

\subsection{Details about the Controlled Study}
\begin{figure}[h]
    \centering
    \includegraphics[width=\columnwidth]{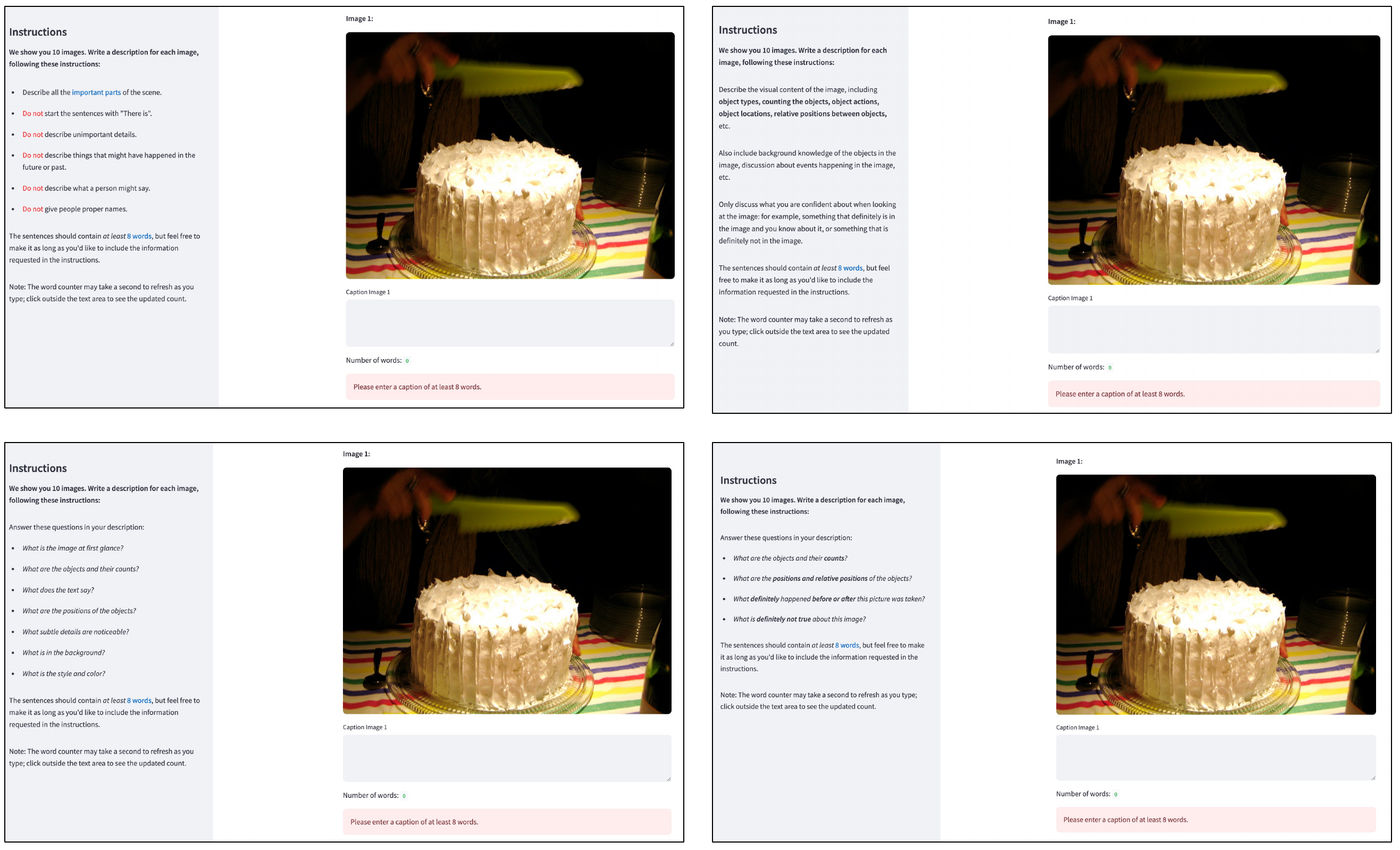}
    \caption{Instructions provided for the COCO (top left), LLaVA-1.5 (top right), PixMo (bottom left) and our (bottom right) sets of instructions.}
    \label{fig:instructions}
\end{figure}

\paragraph{Instructions provided.} The instructions provided to annotators are kept as close as possible to the original papers, with the reasoning-related words kept verbatim. Instructions are visible to the crowdworkers as they scroll through the images they annotate, as shown in Figure \ref{fig:instructions}.

\paragraph{Length experiment.} We study whether asking annotators to write longer captions increases the types of reasoning represented. We collect an additional 50 captions of the first 50 COCO images from our study, with the same instructions as COCO captions. However, we require here that the captions are all at least 50 words. In these 50 captions, 10 have spatial reasoning, 25 have counting, and none have negations/temporal reasoning. The prevalence of spatial and counting is about double that of the study with an 8-word minimum. It is clear that increasing the caption length does encourage some types of reasoning, but it does not serve as a solution to increasing representation of all types of reasoning.

\paragraph{Counting.} We see that the majority of object counts are the number 2, which is easy for annotators to count. However, upon closer inspection of the data, we also see that there are simply fewer images with >2 instances of any given object. This highlights the need to study reporting bias in the image space as well, rather than the text space alone, as discussed in Section \ref{sec:conclusion}.

\subsection{Qualitative Observations}

\paragraph{CLIP ignores negations.} When evaluating negations, we observe that CLIP's performance on negated attributes $\approx$ 100 -- attribute recognition performance. To investigate, we evaluate object negation, and find that CLIP's performance on negated objects $\approx$ 100 -- object recognition performance: the data points on which CLIP gets the negated attribute/object correct are those on which it gets the attribute/object incorrect; showing that the model completely ignores the negation.

\paragraph{Models can count to smaller numbers better.} When evaluating counting, we observe that contrastive and generative VLMs both perform better when counting small numbers than when counting large ones; which also correlates with the numbers' appearance in the training data: annotators are more likely to count smaller numbers of objects---as the number increases, they default to approximations such as ``group of'' and ``several''. 

\paragraph{``Left'' and ``right'' are the most difficult spatial relations for VLMs.} Both contrastive and generative models struggle more with ``left'' and ``right'' than with ``on'' and ``under''. This also correlates with the relations' appearance in the training data, and validates our earlier hypotheses: due to the inherent ambiguity in these two relations (``left'' from which perspective?), symmetric relations like ``next to'' are preferred over asymmetric types of grounding by annotators.

\paragraph{Contrastive VLMs can ignore keywords even when they do occur in the training data.} We show that the phenomena we study are included rarely in captions. When they \textit{are} included, though, it tends to be after the most salient information of the image is already captured by the caption, i.e., they are included as a ``least significant bit'' of information. As such, the contrastive loss allows the model to ignore these parts of the caption completely, as the salient image features are sufficient to retrieve the image in the batch. 

\end{document}